\pdfoutput=1

\documentclass[11pt]{article}

\usepackage{acl}

\usepackage{times}
\usepackage{latexsym}

\usepackage[T1]{fontenc}

\usepackage[utf8]{inputenc}

\usepackage{microtype}

\usepackage{inconsolata}

\usepackage{graphicx}

\newcommand{\rparagraph}[1]{\vspace{1.4mm}\noindent\textbf{#1.}}
\newcommand{\rrparagraph}[1]{\vspace{1.2mm}\noindent\textit{#1.}}
\newcommand{\sparagraph}[1]{\vspace{0.0mm}\noindent\textbf{#1.}}

\newcommand{\ttr}{{\texttt{T-Train}}\xspace}
\newcommand{\tts}{{\texttt{T-Test}}\xspace}

\newcommand{\mlm}{mLM\xspace}
\newcommand{\mlms}{mLMs\xspace}

\newcommand{\xlt}{XLT\xspace}

\newcommand{\wa}{WA\xspace}
\newcommand{\was}{WAs\xspace}
\newcommand{\nllbwa}{TransAlign\xspace}
\newcommand{\labsewa }{AccAlign\xspace}
\newcommand{\mbertwa}{AwsmAlign\xspace}
\newcommand{\tnllbwa}{TransAlign}
\newcommand{\tlabsewa }{AccAlign}
\newcommand{\tmbertwa}{AwsmAlign}
\newcommand{\codec}{Codec\xspace}
\newcommand{\codect}{Codec\xspace}

\usepackage{multirow}
\usepackage{adjustbox}
\usepackage{booktabs}
\usepackage{tabularx}
\usepackage{verbatim}
\usepackage{hyperref}
\usepackage{xspace}
\usepackage{caption}
\usepackage{subcaption}
\usepackage{amssymb}
\usepackage{amsmath}

\title{\nllbwa: Machine Translation Encoders are Strong Word Aligners, Too}

\author{Benedikt Ebing, Christian Goldschmied, and Goran Glava\v{s} \\
  University of W\"{u}rzburg \\ Center for Artificial Intelligence and Data Science (CAIDAS) \\
  \texttt{\{benedikt.ebing, goran.glavas\}@uni-wuerzburg.de} \\}

\begin{document}
\maketitle

\begin{abstract}
In the absence of sizable training data for most world languages and NLP tasks, translation-based strategies such as \textit{translate-test}---evaluating on noisy source language data translated from the target language---and \textit{translate-train}---training on noisy target language data translated from the source language---have been established as competitive approaches for cross-lingual transfer (\xlt). 
For token classification tasks, these strategies require \textit{label projection}: mapping the labels from each token in the original sentence to its counterpart(s) in the translation.
To this end, it is common to leverage multilingual word aligners (\was) derived from encoder language models such as mBERT or LaBSE. Despite obvious associations between machine translation (MT) and \wa, research on extracting alignments with MT models is largely limited to exploiting cross-attention in encoder-decoder architectures, yielding poor \wa results. 
In this work, in contrast, we propose \nllbwa, a novel word aligner that utilizes the encoder of a massively multilingual MT model. We show that \nllbwa not only achieves strong \wa performance but substantially outperforms popular \was and state-of-the-art non-\wa-based label projection methods in MT-based \xlt for token classification. 
\end{abstract}
\section{Motivation and Background}

In recent years, multilingual language models (\mlms) have been positioned as the primary tool for cross-lingual transfer (\xlt). 
By fine-tuning on task data in a high-resource source language, \mlms can make predictions in target languages with no (zero-shot \xlt) or limited (few-shot \xlt) labeled examples \cite{wu-dredze-2019-beto, wang2019cross,lauscher-etal-2020-zero, schmidt-etal-2022-dont}.
However, for \textit{token classification tasks} (e.g., named entity recognition), translation-based \xlt strategies, where a machine translation (MT) model is used to either (1) translate the original target language instance into the (noisy) source language before inference, known as \textit{translate-test} (\tts), or (2) generate noisy target language data by translating the original source language data before training, known as \textit{translate-train} (\ttr)  \cite{pmlr-v119-hu20b, ruder-etal-2021-xtreme, ebrahimi-etal-2022-americasnli, aggarwal-etal-2022-indicxnli, artetxe-etal-2023-revisiting,ebing-glavas-2024-translate}, substantially outperform zero-shot \xlt \cite{chen-etal-2023-frustratingly, garcia-ferrero-etal-2023-projection, DBLP:conf/iclr/LeCR024, parekh-etal-2024-contextual}, especially for low(er)-resource languages \cite{ebing2025devil}. 


Translation-based \xlt strategies for token classification tasks require the additional step of \textit{label projection}: mapping the labeled spans from the original to the translated sentence. A broad body of work addressed label projection starting from task-specific \cite[inter alia]{duong-etal-2013-simpler, ni-etal-2017-weakly, stengel-eskin-etal-2019-discriminative, eskander-etal-2020-unsupervised, fei-etal-2020-cross} and evolving to task-agnostic methods \cite{chen-etal-2023-frustratingly, garcia-ferrero-etal-2023-projection, DBLP:conf/iclr/LeCR024, parekh-etal-2024-contextual}. While \wa-based label projection \cite{och-ney-2003-systematic, dyer-etal-2013-simple, jalili-sabet-etal-2020-simalign, dou-neubig-2021-word, wang-etal-2022-multilingual}--which projects labels by establishing pairwise alignments between tokens in the original sentence and corresponding tokens in the translated sentence---served as baseline throughout, recent work has rendered it less effective than other label projection strategies \cite{chen-etal-2023-frustratingly, garcia-ferrero-etal-2023-projection, DBLP:conf/iclr/LeCR024, parekh-etal-2024-contextual}. However, \citet{ebing2025devil} show that \wa-based label projection can perform at least on a par with these state-of-the-art projection methods as long as: (i) a robust algorithm for label projection and (ii) a sentence encoder-based \wa model are used.

In this work, we hypothesize that multilingual MT models are better suited for producing word alignments than WAs based on multilingual sentence encoders (e.g. LaBSE) \cite{wang-etal-2022-multilingual} or vanilla encoders (e.g. mBERT) \cite{jalili-sabet-etal-2020-simalign, dou-neubig-2021-word}, due to \wa and MT being two highly related and interleaved tasks \cite{och-ney-2003-systematic, callison-burch-etal-2004-statistical, koehn-etal-2007-moses, dyer-etal-2013-simple}.
Yet, research on extracting word alignments from MT models has largely been limited to extracting alignments from the cross attention mechanism, yielding poor \wa performance for transformer-based encoder-decoder MT models \cite{DBLP:journals/corr/BahdanauCB14, ghader-monz-2017-attention, ferrando-costa-jussa-2021-attention-weights}. 

\rparagraph{Contributions}
This is why, \textbf{(1)} we propose \nllbwa, a \wa that leverages (only) the encoder of NLLB \cite{nllbteam2022language}, a massively multilingual encoder-decoder MT model. Next, to its vanilla (non-fine-tuned) variant, we explore the impact of further fine-tuning \nllbwa on parallel \wa data. \textbf{(2)} We extensively evaluate \nllbwa extrinsically on translation-based \xlt for token classification on two established benchmarks covering 28 diverse languages. We find \nllbwa to substantially outperform popular word aligners as well as a state-of-the-art non-\wa-based label projection method. Furthermore, we evaluate \nllbwa intrinsically on the word alignment task showing its strong performance, particularly on words carrying semantic meaning. \textbf{(3)} Finally, we ablate important design decisions including the encoder layer to extract the alignments from and the similarity threshold based on which an alignment is established. We publicly release our code and data in the following repository: \href{https://github.com/bebing93/transalign}{https://github.com/bebing93/transalign}.
\section{An MT Encoder as a Word Aligner}
\label{sec:method}
The task of word alignment aims at finding semantically corresponding pairs of words between a source language sentence $x=(x_1,x_2,...,x_n)$ and target language sentence $y=(y_1,y_2,...,y_m)$: 
\begin{equation}
    A=\{(x_i,y_j): x_i \in x, y_j \in y\}.
\end{equation}

\rparagraph{Extracting Alignments}
For \nllbwa, we extract word alignments from the contextualized embeddings produced by the \textit{encoder} of a multilingual encoder-decoder MT model. We separately feed the source language sentence $x$ and target language sentence $y$ through the encoder, obtaining their contextualized representations $h_x$ and $h_y$, respectively. Following prior work \cite{jalili-sabet-etal-2020-simalign, dou-neubig-2021-word, wang-etal-2022-multilingual}, we next obtain the token similarity matrix $S_{xy}$:
\begin{equation}
    S_{xy}=h_xh_y^T
\end{equation}
We row- and column-normalize the similarity matrix using softmax to obtain $\hat{S}_{xy}$ and $\hat{S}_{yx}$, which capture the similarity from $x$ to $y$ and $y$ to $x$. Finally, we compute the alignment matrix $A$ by intersecting the two similarity matrices:
\begin{equation}
    A_{ij} = \begin{cases}
        1 & \text{if } (\hat{S}_{xy})_{ij} > c \text{ and } (\hat{S}_{yx})_{ji} > c \\
        0 & \text{otherwise}
\end{cases}
\end{equation}
where $c$ is the alignment threshold and $A_{ij}=1$ indicates that two tokens are aligned. As the MT encoder operates on the level of sub-word tokens, we consider two words to be aligned if any of their sub-word tokens are aligned, in line with the prior \wa work \cite{jalili-sabet-etal-2020-simalign, dou-neubig-2021-word, wang-etal-2022-multilingual}.

\rparagraph{Fine-Tuning for Word Alignment}
Additionally, we explore fine-tuning \nllbwa on a word alignment-specific objective to further improve performance. Different from related work---that employed full fine-tuning \cite{dou-neubig-2021-word, wang-etal-2022-multilingual} or adapter-based fine-tuning \cite{ wang-etal-2022-multilingual}---we opt for LoRA \cite{DBLP:conf/iclr/HuSWALWWC22} as it does not increase model depth while maintaining parameter efficiency.
We resort to the following loss function for \wa fine-tuning:
\begin{equation}
    L = \sum_{ij}{\hat{A}_{ij}{\frac{1}{2}}(\frac{(\hat{S}_{xy})_{ij}}{n} + \frac{(\hat{S}_{yx})_{ji}}{m})},
\end{equation}
where $\hat{A}$ refers to the gold alignments from the labeled data and $n$ (or $m$) is the number of tokens in sentence $x$ (or $y$)  \cite{dou-neubig-2021-word, wang-etal-2022-multilingual}.

\section{Experiments}

With label projection as the key remaining application of \was, we first evaluate \nllbwa on translation-based \xlt for token classification. We then additionally benchmark \nllbwa intrinsically on word alignment itself.  

\subsection{Experimental Setup}

\sparagraph{Extrinsic Evaluation}
We evaluate \tts\footnote{\tts is shown to outperform \ttr \cite{DBLP:conf/iclr/LeCR024, ebing2025devil}.} as our translation-based \xlt strategy: at inference, the original target language sentence is translated into English. We then gather predictions for the translated English sentence from a fine-tuned downstream \mlm. Next, we use a WA to extract pairwise alignments between the tokens in the translated English sentence and the original target language sentence. Having obtained the alignments, we follow the span-based label projection algorithm of \citet{ebing2025devil}\footnote{The algorithm projects labels across spans and not individual tokens, and can compensate for some word alignment errors. For details, we refer the reader to the original work.} to map the predictions back to the target language sentence. We compare \nllbwa against two popular \was: (i) \mbertwa \cite{dou-neubig-2021-word}, based on multilingual BERT, and (ii) \labsewa \cite{wang-etal-2022-multilingual}, based on the multilingual sentence encoder LaBSE. We utilize the word alignment-specific fine-tuned variants of all WAs. For a fair comparison between WAs, we follow the same fine-tuning protocol and extract alignments following the same procedure as described in section \ref{sec:method}. Additionally, we benchmark \nllbwa against \codec \cite{DBLP:conf/iclr/LeCR024}: a state-of-the-art non-\wa-based label projection method that projects the predictions back to the original target language sentence by means of constrained decoding. We evaluate for 28 diverse languages on two established token classification tasks: named entity recognition (NER) and slot labeling (SL). For NER, we utilize MasakhaNER2.0 (Masakha) \cite{adelani-etal-2022-masakhaner}, which encompasses low-resource languages from Sub-Saharan Africa. For SL, the evaluation dataset is xSID \cite{van-der-goot-etal-2021-masked}, covering mid- to high-resource languages and dialects. For our downstream \mlms, we fine-tune XLM-R Large \cite{conneau-etal-2020-unsupervised} and DeBERTaV3 Large \cite{he2023debertav3improvingdebertausing} on the English portion of our data. We run experiments with 3 random seeds and report the mean F\textsubscript{1} score and standard deviation. We provide details for the extrinsic evaluation in Appendix \ref{app:expdetails}.

\rparagraph{Intrinsic Evaluation}
We evaluate \nllbwa on 8 language pairs: en-cz/de/fr/hi/ja/ro/sv/zh and compare it against the same \wa baselines. All \was are evaluated in their non-fine-tuned variant. 
We report AER for each language pair. 
We provide details of the intrinsic evaluation in Appendix \ref{app:exp_in_details}. 

\rparagraph{\nllbwa}
For both extrinsic and intrinsic evaluation, we use the encoder of the distilled 600M parameter version of NLLB \cite{nllbteam2022language} as the backbone of \nllbwa. We extract alignments after the last (i.e., 12th) layer using an alignment threshold of $c=0.001$.

\subsection{Main Results}
\rparagraph{Extrinsic Evaluation}
\begin{table}[]
\small
\setlength{\tabcolsep}{8.1pt}
\centering
\begin{tabular}{@{}lcccc@{}}
\toprule
             &      & Masakha    & xSID    & Avg     \\ \midrule
ZS           & X    & $52.9_{\pm1.8}$  & $76.5_{\pm1.4}$   & $64.7_{\pm1.7}$    \\ \midrule
\multicolumn{5}{c}{\textit{\textbf{Translate-Test: non-WA}}} \\ \midrule
\codect        & X    & $72.0_{\pm0.5}$  & $80.1_{\pm0.3}$    & $76.1_{\pm0.4}$    \\
\codect        & D    & $72.4_{\pm0.4}$  & $80.2_{\pm0.4}$    & $76.3_{\pm0.4}$    \\ \midrule
\multicolumn{5}{c}{\textit{\textbf{Translate-Test: WA}}} \\ \midrule
\tmbertwa  & X    & $68.4_{\pm0.4}$       & $78.8_{\pm0.3}$    & $73.6_{\pm0.4}$    \\
\tmbertwa  & D    & $68.8_{\pm0.4}$       & $78.7_{\pm0.4}$    & $73.8_{\pm0.4}$    \\
\tlabsewa  & X    & $72.3_{\pm0.4}$       & $80.9_{\pm0.3}$    & $76.6_{\pm0.4}$    \\
\tlabsewa  & D    & $72.7_{\pm0.4}$       & $80.8_{\pm0.4}$    & $76.8_{\pm0.4}$    \\
\tnllbwa   & X    & $73.9_{\pm0.4}$       & $\mathbf{82.2}_{\pm0.4}$    & $78.1_{\pm0.4}$    \\
\tnllbwa   & D    & $\mathbf{74.3}_{\pm0.4}$       & $\mathbf{82.2}_{\pm0.4}$    & $\mathbf{78.3}_{\pm0.4}$    \\ \bottomrule
\end{tabular}
\caption{Main results for translation-based XLT for token classification. Results with XLM-R (X) and DeBERTa (D). All \wa models are evaluated in their fine-tuned variant. We report mean F1.}
\label{tab:main_downstream}
\end{table}
%
Table \ref{tab:main_downstream} outlines the \tts results for the fine-tuned \was and \codec. We demonstrate that all \tts strategies exceed zero-shot \xlt substantially reaching an improvement of up to 13.4\% on average (with \nllbwa and XLM-R). Comparing \nllbwa against the other \wa baselines, we find it to clearly outperform \mbertwa and \labsewa by 5.5\% and 1.5\% on average.\footnote{Since \nllbwa covers substantially more languages than \labsewa, we provide additional experiments demonstrating that the improved performance does not stem from broader language coverage (see Appendix \ref{app:supported_langs}).}
Not only does \nllbwa outperform popular \was in translation-based \xlt for token classification, but it also improves over the competitive non-\wa-based label projection method \codec by 2\% on average.
This finding is noteworthy as \nllbwa is a \textit{fair} baseline for \codec: both approaches use a fine-tuned NLLB model of the same size for label projection. However, \nllbwa is computationally more efficient as it only uses the encoder of NLLB and thus avoids the costly constrained decoding of \codec \cite{DBLP:conf/iclr/LeCR024}.



\rparagraph{Intrinsic Evaluation}
\begin{table}[]
\small
\setlength{\tabcolsep}{1.3pt}
\centering
\begin{tabular}{@{}lcccccccc@{}}
\toprule
              & en-zh & en-cs & en-fr & en-de & en-hi & en-ja & en-ro & en-sv \\ \midrule
\multicolumn{9}{c}{\textit{\textbf{All Words}}} \\ \midrule
\mbertwa & 18.2 & 12.3 & 6.3  & 18.6 & 42.9 & 46.2 & 28.9 & 9.9  \\
\labsewa & \textbf{16.2} & 9.3  & \textbf{5.2}  & \textbf{16.4} & 30.4 & 43.3 & 20.8 & \textbf{7.3}  \\
\nllbwa  & 18.8 & \textbf{8.9}  & 6.8  & 17.7 & \textbf{29.4} & \textbf{43.2} & \textbf{20.6} & 7.8 \\ \midrule
\multicolumn{9}{c}{\textit{\textbf{w/o Stopwords}}} \\ \midrule
\mbertwa & 12.5 & 10.6 & 5.3 & 14.2 & 35.6 & \textbf{35.3} & 22.0 & 9.2  \\
\labsewa & 10.7 & 6.8 & 4.3 & \textbf{11.6} & 24.9 & 37.5 & 16.1 & 5.8  \\
\nllbwa  & \textbf{10.6} & \textbf{6.3} & \textbf{4.0} & 11.8 & \textbf{23.4} & 36.5 & \textbf{15.2} & \textbf{5.1} \\
\bottomrule
\end{tabular}
\caption{Main results for word alignment evaluation. All \wa models are evaluated in their vanilla (non-fine-tuned) variant. We report the AER considering all words and without considering stopwords.}
\label{tab:main_wa}
\vspace{-0.5em}
\end{table}
We present the results for intrinsic evaluation in Table \ref{tab:main_wa}. Considering all words in the source and target sentence equally, we find that \nllbwa produces the best results for 4 out of 8 language pairs (\labsewa reaches the best performance on the remaining ones). While \nllbwa and \labsewa perform similarly on alignment itself, our \nllbwa exhibited stronger downstream \xlt performance (Table \ref{tab:main_downstream}). 
For example, in intrinsic evaluation, \labsewa outperforms \nllbwa for Chinese and German by 2.6\% and 1.3\%, respectively. In contrast, for \tts on xSID (see App.~\ref{app:detailed_main}), the trend turns around: \nllbwa outperforms \labsewa for both Chinese (0.7\%) and German (2.2\%). 

These results point to a mismatch between the standard word alignment evaluation that treats each word in the input as equally important and label projection for translation-based \xlt that requires correct alignments on a subset of the input sentence. 
Commonly, labeled spans in downstream evaluation span words that carry meaning (e.g., named entities). We thus additionally report the alignment results by excluding stopwords---words with little semantic meaning---from the evaluation. Results presented in Table \ref{tab:main_wa} (bottom half) support our hypothesis: not accounting for the (accuracy of) stopword alignment, \nllbwa outperforms both baselines consistently: this means it produces more accurate alignments between content words, which explains why it yields downstream \xlt gains. 


\subsection{Analysis}

\begin{figure}[t!]
\centering
\includegraphics{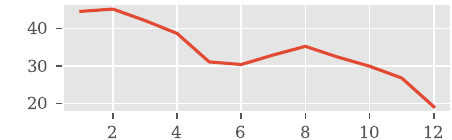}
\caption{Word alignment performance across layers of vanilla \nllbwa. We present the average AER over all 8 language pairs.}
\label{fig:layer}
\vspace{-0.5em}
\end{figure}

\rparagraph{Performance per Layer}
The layer from which we extract the alignments can have a substantial impact on performance \cite{jalili-sabet-etal-2020-simalign, dou-neubig-2021-word, wang-etal-2022-multilingual}. Figure \ref{fig:layer} shows the average AER performance for all layers of vanilla \nllbwa: using the last layer of \nllbwa substantially outperforms using any other layer. 

\rparagraph{Alignment Threshold}
\begin{figure}[t!]
\centering
\includegraphics{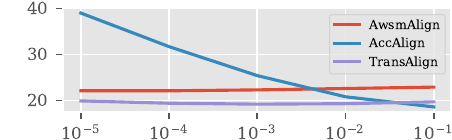}
\caption{Word alignment performance for different thresholds of $c$. We evaluate vanilla \was and present the average AER over all 8 language pairs.}
\label{fig:softmax}
\vspace{-0.5em}
\end{figure}
The threshold parameter $c$ decides whether two tokens are considered to be aligned. We ablate the choice of $c$ for all \was in their vanilla variant (see Figure \ref{fig:softmax}). 
While \mbertwa and \nllbwa are robust to the threshold value, we find \labsewa's performance to severely vary with the value of $c$. 


\rparagraph{Impact of \wa Fine-Tuning}
We obtained the best results for our \wa-fine-tuned \nllbwa (Table \ref{tab:main_downstream}). We next assess the contribution of word alignment fine-tuning for all \was on downstream MT-based \xlt performance (see Table \ref{tab:impact_ft}). We find that fine-tuning improves the \xlt results for all \was, but the gains are more pronounced for \was with weaker initial performance: \mbertwa improves by 3.6\% compared to 0.6\% for \nllbwa. We also note that using a stronger \wa model is more beneficial than fine-tuning: vanilla \nllbwa outperforms the \wa-fine-tuned \labsewa by 0.7\%. 

\begin{table}[ht!]
\small
\setlength{\tabcolsep}{12pt}
\centering
\begin{tabular}{@{}lccc@{}}
\toprule
              & Masakha & xSID & Avg  \\ \midrule
\multicolumn{4}{c}{\textit{\textbf{Non-Fine-Tuned \was}}} \\ \midrule
\mbertwa & $66.2_{\pm0.3}$    & $74.1_{\pm0.3}$ & $70.2_{\pm0.3}$ \\
\labsewa & $71.2_{\pm0.4}$    & $80.0_{\pm0.4}$ & $75.6_{\pm0.4}$ \\
\nllbwa  & $73.5_{\pm0.4}$    & $81.8_{\pm0.4}$ & $77.7_{\pm0.4}$ \\ \midrule
\multicolumn{4}{c}{\textit{\textbf{Fine-Tuned \was}}} \\ \midrule
\mbertwa      & $68.8_{\pm0.4}$       & $78.7_{\pm0.4}$    & $73.8_{\pm0.4}$ \\
\labsewa      & $72.7_{\pm0.4}$       & $80.8_{\pm0.4}$    & $76.8_{\pm0.4}$ \\
\nllbwa       & $\mathbf{74.3}_{\pm0.4}$       & $\mathbf{82.2}_{\pm0.4}$    & $\mathbf{78.3}_{\pm0.4}$ \\ \bottomrule
\end{tabular}
\caption{Impact of \wa fine-tuning on translation-based XLT for token classification. Results with DeBERTa. 
}
\label{tab:impact_ft}
\vspace{-0.5em}
\end{table}
\section{Conclusion}

In this work, we proposed \nllbwa, a new word aligner (\wa) that leverages the encoder of NLLB, a massively multilingual encoder-decoder MT model. Our extrinsic evaluation on translation-based \xlt for token classification on two established benchmarks covering 28 languages, shows that \nllbwa outperforms popular existing \was as well as state-of-the-art non-\wa-based label projection methods. Furthermore, our intrinsic word alignment evaluation reveals that, \nllbwa aligns content words (rather than functional words) in particular better than existing \was, which then reflects in downstream \xlt gains. 
\section{Limitations}
We focused on choosing well-established and representative tasks for token classification. However, in NLP, multilingual evaluation benchmarks are often created by translating the data from an existing high-resource language followed by post-editing. This applies to xSID and some languages of Masakha. As a result, the newly introduced languages might contain translation artifacts referred to as \textit{translationese}. Prior work \cite{artetxe-etal-2020-translation, artetxe-etal-2023-revisiting} stated that translation-based \xlt strategies might lead to exploitation of translationese, slightly overestimating performance.

Our intrinsic evaluation points to a potential mismatch between the word alignment task and the extrinsic evaluation on translation-based \xlt for token classification. Our results suggest that the mismatch stems from the discrepancy of treating all words equally (intrinsic evaluation) against focusing on a specific subset of words (extrinsic evaluation). While we hypothesize as to why MT models perform worse in aligning words with little semantic meaning than sentence encoders, further work is needed to test our hypothesis.

\section*{Acknowledgments}
Simulations were performed with computing resources from Julia 2. Julia 2 was funded as DFG project as “Forschungsgroßgerät nach Art 91b GG” under INST 93/1145-1 FUGG". Further, simulations were performed with computing resources granted by WestAI under project westai8850.  

\bibliography{anthology,custom}

\appendix
\section{Experimental Details: Extrinsic Evaluation}
\label{app:expdetails}
\sparagraph{Machine Translation} For translation, we utilize the state-of-the-art massively multilingual NLLB model with 3.3B parameters \cite{nllbteam2022language}. Following prior work \cite{artetxe-etal-2023-revisiting, ebing-glavas-2024-translate, ebing2025devil}, we decode using beam search with a beam size of $5$. For Masakha \cite{adelani-etal-2022-masakhaner} and xSID \cite{van-der-goot-etal-2021-masked}, we concatenated the pre-tokenized input on whitespace before translation. We deviate from this for the Chinese data in xSID, where we merge Chinese tokens without whitespace. Additionally, the dialect \textit{South Tyrol} (de-st) in xSID is not supported by NLLB. We translate the dialect pretending it to be German (i.e., using the German language code) as it is closely related to the latter. We accessed all datasets through the Hugging Face library and ensured compliance with the licenses. All translations were run on a single A100 with 40GB VRAM.

\rparagraph{Word Aligners}
We will publicly release our word alignment code (Apache 2.0 license) and the model checkpoints for the fine-tuned \nllbwa (CC-BY-NC 4.0 license). Next to \nllbwa, we re-implemented two popular word aligners as our baselines: \mbertwa \cite{dou-neubig-2021-word} and \labsewa \cite{wang-etal-2022-multilingual}. We chose the code repository of SimAlign \cite{jalili-sabet-etal-2020-simalign} as the starting point for our implementation. We accessed the code through their repository: (\href{https://github.com/cisnlp/simalign}{https://github.com/cisnlp/simalign}). Following \citet{dou-neubig-2021-word}, we extracted alignments for \mbertwa after the 8th layer using an alignment threshold of $c=0.001$. For \labsewa, we use the 6th layer and an alignment threshold of $c=0.1$ \cite{wang-etal-2022-multilingual}. We comply with the licenses of \mbertwa (BSD 3-Clause) and SimAlign (MIT). We could not find licensing information for \labsewa.  

\rparagraph{Codec}
Codec \cite{DBLP:conf/iclr/LeCR024} is a label projection method that leverages constrained decoding as part of a two-step translation procedure.
In the first step, the source sentence is translated into the target language (e.g., from English: ``This is New York'' to German: ``Das ist New York''). Then, in the second step, tags are inserted around the labeled spans in the source sentence (English: ``This is [ New York ]''). The marked sentence is fed again as input to the MT model: during decoding, the MT model is now constrained to generate only the tokens from the translation obtained in the first step (``Das'', ``ist'', ``New'', ``York'') or a tag (``['', ``]''). We chose \codec as a representative method for non-\wa-based label projection: \citet{ebing2025devil} suggest that \codec performs on par or better than comparable non-\wa-based label projection methods \cite{chen-etal-2023-frustratingly, garcia-ferrero-etal-2023-projection, parekh-etal-2024-contextual}.
To project the labels for \tts, we used the publicly available code repository of \codec: \href{https://github.com/duonglm38/Codec}{https://github.com/duonglm38/Codec}. While an implementation for Masakha is already provided, we extended their implementation to handle label projection for xSID. We adhered to the hyperparameters in their repository and followed the existing implementation closely. 
The constrained decoding (i.e., inserting the tags post-translation) requires a fine-tuned NLLB that is able to preserve/insert tags. Therefore, we follow \citet{DBLP:conf/iclr/LeCR024} using the fine-tuned 600M parameter version of NLLB released by \citet{chen-etal-2023-frustratingly}. We could not find licensing information for Codec. 

\rparagraph{Label Projection} 
We follow the span-based label projection procedure used by \cite{ebing2025devil}. The algorithm projects labels across spans and not individual tokens and can compensate for some word alignment errors. For details, we refer the reader to the original work. Unlike their work, we do not apply filtering heuristics for \tts.

\rparagraph{Word Aligner Fine-Tuning}
For fine-tuning, we apply LoRA to the feed-forward sublayer of each encoder layer. We train each \wa for $20$ epochs using a learning rate of $1e^{-4}$. The rank is set to $8$ and alpha to $32$. We apply LoRA dropout with $0.01$. For \wa training, we utilize the labeled data from the intrinsic evaluation (see Table \ref{tab:eval_data}).

\rparagraph{Downstream Fine-Tuning}
We train both tasks (NER and SL) for $10$ epochs using an effective batch size of $32$. In case we can not fit the desired batch size, we utilize gradient accumulation. The learning rate is set to $1e^{-5}$ with a weight decay of $0.01$. We implement a linear schedule of $10\%$ warm-up and employ mixed precision. We evaluate models at the last checkpoint of training. We use the seqeval F1 implementation accessed through the Hugging Face library. Further, we access our downstream models---XLM-RoBERTa Large and DeBERTaV3 Large---through the Hugging Face library. All downstream training and evaluation runs were completed on a single V100 with $32$GB VRAM. We estimate the GPU time to be $2000$ hours across all translations and downstream fine-tunings.

\rparagraph{Datasets}

\rrparagraph{MasakhaNER2.0} Our experiments cover $18$ out of $20$ languages that are supported by NLLB. Note that Google Translate (GT) does not support all $18$ languages. Following, we mark the $11$ languages that are supported by GT with an additional asterisk:  Bambara (bam)*, Ewé (ewe)*, Fon (fon), Hausa (hau)*, Igbo (ibo)*, Kinyarwanda (kin)*, Luganda (lug), Luo (luo), Mossi (most), Chichewa (nya), chiShona (sna)*, Kiswahili (saw)*, Setswana (tsn), Akan/Twi (twi)*, Wolof (wol), isiXhosa (xho)*, Yorùrbá (yor)*, and isiZulu (zul)*. As source data, we use the English training (14k instances) and validation portions (3250 instances) of CoNLL \cite{tjong-kim-sang-de-meulder-2003-introduction}.

\rrparagraph{xSID} We evaluate $10$ languages all covered by NLLB and GT: Arabic (ar), Danish (da), German (de), South-Tyrolean (de-st), Indonesian (id), Italian (it), Kazakh (kk), Dutch (nl), Turkish (tr), and Chinese (zh). Following \citet{razumovskaia-etal-2023-transfer}, we excluded Japanese from the evaluation because it only has half of the validation and test instances and spans only a fraction of entities compared to the other languages. Moreover, we exclude Serbian as the evaluation data is written in the Latin script whereas NLLB was only trained in the Cyrillic script. xSID is an evaluation-only dataset. Therefore, we follow \citet{van-der-goot-etal-2021-masked} and use their publicly released English data for training and validation. The instances are sourced from the Snips \cite{coucke2018snipsvoiceplatformembedded} and Facebook \cite{schuster-etal-2019-cross-lingual} SL datasets. We deduplicate the training instances, ending up with over 36k training and 300 validation examples.

\section{Experimental Details: Intrinsic Evaluation}
\label{app:exp_in_details}
\sparagraph{Word Alignment Baselines}
We use the same \wa models as for the extrinsic evaluation---\mbertwa and \labsewa (see App.~\ref{app:expdetails}). All \was are evaluated in their non-fine-tuned variant.

\rparagraph{Languages} We evaluate the following 8 language pairs: English-Chinese (en-zh), English-Czech (en-cz), English-French (en-fr), English-German (en-de), English-Hindi (en-hi), English-Japanese (en-ja), English-Romanian (en-ro) and English-Swedish (en-sv). We provide details on the used datasets in Table \ref{tab:eval_data}.

\rparagraph{Stopword Filtering}
For the results in Table \ref{tab:main_wa}, we applied stopword filtering prior to AER computation. We identified stopwords from the English source sentences using the stopword list provided by NLTK \cite{elhadad-2010-book} and removed corresponding target language words accordingly. The NLTK source code is published under the Apache 2.0 license. We comply with their license.

\section{Further Analysis: Robustness of Fine-Tuning}
\begin{figure}[ht!]
\centering
\includegraphics{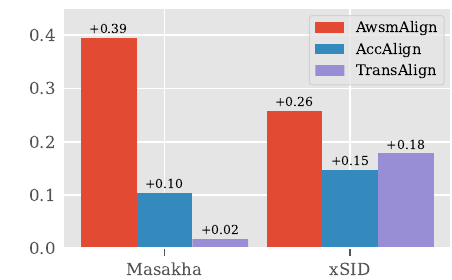}
\caption{Variance of  WA model fine-tuning with three distinct random seeds evaluated on translation-based XLT. Results with DeBERTa.}
\label{fig:robustness_ft}
\vspace{-0.5em}
\end{figure}

For the application of a fine-tuned \wa model, only a single seed of a fine-tuned model will eventually be used. Therefore, we ablate the variance of the random seed chosen for fine-tuning. We fine-tune \mbertwa, \labsewa, and \nllbwa on three distinct random seeds and evaluate them on translation-based \xlt. The resulting variance is depicted in Figure \ref{fig:robustness_ft}. We observe little impact by the choice of the random seed for \nllbwa: for xSID the variance is comparable to that of \mbertwa and \labsewa, while for Masakha, it is substantially lower.

\section{Further Analysis: MT Model}

In translation-based XLT for token classification, it is pragmatic to use the encoder of the MT model for word alignment since (i) only a single model is required for the label projection pipeline (i.e., translation and label projection) and (ii) the language coverage of target languages is ensured for both steps. However, open access to the encoder of the MT model is required. With closed commercial MT models being considered to produce superior translation quality, we explore whether the gains obtained by \nllbwa are orthogonal to the MT model. Our results in Table \ref{tab:translations} suggest that \nllbwa does not depend on its \textit{own} translations. The performance improvements obtained by label projection with \nllbwa are orthogonal to gains obtained by higher translation quality.

\begin{table}[t!]
\small
\setlength{\tabcolsep}{6.5pt}
\centering
\begin{tabular}{@{}llccc@{}}
\toprule
         &        & Masakha & xSID & Avg  \\ \midrule
\mbertwa & NLLB   & $69.2_{\pm0.4}$    & $78.7_{\pm0.4}$ & $74.0_{\pm0.4}$ \\
\labsewa & NLLB   & $73.7_{\pm0.4}$    & $80.8_{\pm0.4}$ & $77.3_{\pm0.4}$ \\
\nllbwa  & NLLB   & $75.1_{\pm0.5}$    & $82.2_{\pm0.4}$ & $78.7_{\pm0.5}$ \\ \midrule
\mbertwa & GT & $70.6_{\pm0.3}$    & $80.1_{\pm0.4}$ & $75.3_{\pm0.4}$ \\
\labsewa & GT & $75.2_{\pm0.4}$    & $82.1_{\pm0.4}$ & $78.6_{\pm0.4}$ \\
\nllbwa  & GT & $76.4_{\pm0.5}$    & $83.6_{\pm0.4}$ & $80.0_{\pm0.4}$ \\ \bottomrule
\end{tabular}
\caption{Results for translation-based \xlt for token-level tasks with translations obtained from different MT models---Google Translation (GT) and NLLB (NLLB). Results with DeBERTa. 
}
\label{tab:translations}
\vspace{-0.5em}
\end{table}

\section{Further Analysis: Language Coverage}
\label{app:supported_langs}
NLLB has seen substantially more languages in pretraining than LaBSE (200 vs. 109 languages). To ensure that performance improvements obtained by \nllbwa do not simply stem from broader language coverage, we evaluate \nllbwa and \labsewa on a subset of languages seen in the pretraining of both models. We observe that \nllbwa still outperforms \labsewa even on a subset of languages seen by both models (see Table \ref{tab:supported_langs}).

\begin{table}[t!]
\small
\setlength{\tabcolsep}{13.3pt}
\centering
\begin{tabular}{@{}lccc@{}}
\toprule
         & Masakha & xSID & Avg  \\ \midrule
\labsewa & $74.1_{\pm0.5}$    & $83.2_{\pm0.4}$ & $78.7_{\pm0.4}$ \\
\nllbwa  & $75.4_{\pm0.5}$    & $84.7_{\pm0.4}$ & $80.0_{\pm0.4}$ \\ \bottomrule
\end{tabular}
\caption{Results for translation-based \xlt for token-level tasks only evaluating languages seen in the pretraining of both \was. Results with DeBERTa. 
}
\label{tab:supported_langs}
\vspace{-0.5em}
\end{table}

\begin{table}[t!]
\small
\setlength{\tabcolsep}{7pt}
\centering
\begin{tabular}{@{}llccc@{}}
\toprule
           &  & Masakha & xSID & Avg  \\ \midrule
\nllbwa & 600M & $74.3_{\pm0.4}$       & $82.2_{\pm0.4}$    & $78.3_{\pm0.4}$ \\
\nllbwa & 3.3B & $74.5_{\pm0.4}$    & $81.4_{\pm0.4}$ & $78.0_{\pm0.4}$ \\ \bottomrule
\end{tabular}
\caption{Results for translation-based \xlt for token-level tasks with different sizes of NLLB as \wa. Results with DeBERTa. 
}
\label{tab:nllb_size}
\end{table}

\section{Further Analysis: NLLB Model Size}
NLLB is released in different model sizes ranging from 600M up to 54B parameters. Table \ref{tab:nllb_size} compares the fine-tuned \nllbwa in two different model sizes. We evaluate the 600M (distilled) and 3.3B parameter models on translation-based \xlt for token classification. Our results reveal that the larger model does not provide any advantage. Hence, we used the 600M parameter model for our main results.

\begin{table*}[]
\scriptsize
\setlength{\tabcolsep}{15.4pt}
\centering
\begin{tabular}{@{}lllc@{}}
\toprule
Lang  & Source & Link & \#Sents \\ \midrule
en-zh &  \cite{10.5555/2886521.2886640}      &   \href{https://nlp.csai.tsinghua.edu.cn/~ly/systems/TsinghuaAligner/TsinghuaAligner.html}{https://nlp.csai.tsinghua.edu.cn/~ly/systems/TsinghuaAligner/TsinghuaAligner.html}   & 450    \\
en-cs &  \cite{marecek-etal-2008-automatic}      &  \href{https://ufal.mff.cuni.cz/czech-english-manual-word-alignment}{https://ufal.mff.cuni.cz/czech-english-manual-word-alignment}   & 2400   \\
en-fr &  \cite{mihalcea-pedersen-2003-evaluation}      & \href{https://web.eecs.umich.edu/~mihalcea/wpt/}{https://web.eecs.umich.edu/~mihalcea/wpt/}    & 447    \\
en-de &  \cite{vilar-etal-2006-aer}      & \href{https://www-i6.informatik.rwth-aachen.de/goldAlignment/}{https://www-i6.informatik.rwth-aachen.de/goldAlignment/}     & 508    \\
en-hi &  \cite{aswani-gaizauskas-2005-aligning}      &  \href{https://web.eecs.umich.edu/~mihalcea/wpt05/}{https://web.eecs.umich.edu/~mihalcea/wpt05/}    & 90     \\
en-ja &  \cite{neubig11kftt}      &  \href{https://www.phontron.com/kftt/}{https://www.phontron.com/kftt/}     & 582    \\
en-ro &  \cite{mihalcea-pedersen-2003-evaluation}     &  \href{https://web.eecs.umich.edu/~mihalcea/wpt05/}{https://web.eecs.umich.edu/~mihalcea/wpt05/}     & 248    \\
en-sv &  \cite{holmqvist-ahrenberg-2011-gold}      & \href{https://www.ida.liu.se/divisions/hcs/nlplab/resources/ges/}{https://www.ida.liu.se/divisions/hcs/nlplab/resources/ges/}     & 192    \\ \midrule
en-nl & \cite{DBLP:conf/lrec/Macken10} & \href{http://www.tst.inl.nl/}{http://www.tst.inl.nl/} & 372 \\
en-tr & \cite{cakmak-etal-2012-word} & \href{https://web.itu.edu.tr/gulsenc/resources.htm}{https://web.itu.edu.tr/gulsenc/resources.htm} & 300 \\
en-es & \cite{graca-etal-2008-building} & \href{https://www.hlt.inesc-id.pt/w/Word_Alignments}{https://www.hlt.inesc-id.pt/w/Word\_Alignments} & 100 \\
en-pt & \cite{graca-etal-2008-building} & \href{https://www.hlt.inesc-id.pt/w/Word_Alignments}{https://www.hlt.inesc-id.pt/w/Word\_Alignments}  & 100 \\
\bottomrule
\end{tabular}
\caption{Datasets used for intrinsic evaluation and fine-tuning of \was. The upper half is used for intrinsic evaluation and \wa fine-tuning, whereas the lower half is only used for \wa fine-tuning. For the fine-tuning, we held out 100 randomly selected instances of the en-cs dataset as validation portion.}
\label{tab:eval_data}
\vspace{-0.5em}
\end{table*}

\begin{table*}
\section{Detailed Results: Main Results}
\label{app:detailed_main}
\scriptsize
\setlength{\tabcolsep}{4.5pt}
\centering
\begin{tabular}{@{}llccccccccccccccccccc@{}}
\toprule
           &   & bam  & ewe  & fon  & hau  & ibo  & kin  & lug  & luo  & mos  & nya  & sna  & swa  & tsn  & twi  & wol  & xho  & yor  & zul  & Avg  \\ \midrule
ZS         & X  & 43.4 & 72.8 & 61.0 & 73.5 & 49.9 & 46.3 & 64.9 & 55.0 & 56.1 & 51.1 & 34.4 & 88.1 & 51.5 & 49.5 & 56.2 & 22.2 & 35.1 & 41.5 & 52.9 \\ \midrule
\multicolumn{21}{c}{\textit{\textbf{Translate-Test: non-WA}}}                                                                                       \\ \midrule
\codec      & X & 54.5 & 78.8 & 67.4 & 72.9 & 72.8 & 77.6 & 83.6 & 72.8 & 49.4 & 78.1 & 79.3 & 82.2 & 79.2 & 72.5 & 67.3 & 72.5 & 58.4 & 77.1 & 72.0 \\
\codec      & D & 54.3 & 79.1 & 68.0 & 73.3 & 73.9 & 78.2 & 83.5 & 74.2 & 48.8 & 79.0 & 79.8 & 82.9 & 79.3 & 73.1 & 67.8 & 72.6 & 58.0 & 77.0 & 72.4 \\ \midrule
\multicolumn{21}{c}{\textit{\textbf{Translate-Test: WA}}}                                                                                           \\ \midrule
\mbertwa  & X & 51.4 & 78.7 & 61.3 & 70.9 & 75.4 & 66.8 & 82.7 & 72.2 & 47.6 & 77.5 & 71.8 & 81.5 & 79.5 & 70.8 & 62.1 & 56.0 & 61.7 & 63.8 & 68.4 \\
\mbertwa  & D & 51.1 & 78.8 & 62.1 & 71.4 & 77.0 & 67.6 & 82.5 & 73.6 & 47.6 & 77.9 & 72.3 & 82.1 & 79.8 & 71.7 & 62.5 & 56.2 & 61.2 & 63.8 & 68.8 \\
\labsewa   & X & 54.4 & 79.9 & 69.7 & 74.7 & 75.2 & 70.8 & 84.4 & 72.6 & 53.1 & 78.6 & 81.7 & 83.0 & 80.0 & 71.2 & 64.9 & 73.2 & 55.4 & 78.7 & 72.3 \\
\labsewa   & D & 53.8 & 79.9 & 70.1 & 75.2 & 76.7 & 71.5 & 84.2 & 74.1 & 53.2 & 79.1 & 82.3 & 83.6 & 80.4 & 72.0 & 65.4 & 73.3 & 55.3 & 78.8 & 72.7 \\
\nllbwa & X & 56.8 & 80.8 & 72.8 & 74.9 & 75.8 & 71.0 & 84.8 & 74.7 & 54.0 & 78.8 & 82.2 & 82.3 & 82.2 & 75.1 & 68.6 & 73.8 & 62.8 & 79.0 & 73.9 \\
\nllbwa & D & 56.6 & 80.8 & 73.3 & 75.4 & 77.3 & 71.7 & 84.6 & 76.3 & 53.7 & 79.2 & 82.8 & 82.9 & 82.6 & 75.8 & 69.3 & 74.0 & 62.5 & 79.1 & 74.3 \\ \bottomrule
\end{tabular}
\caption{Detailed main results for translation-based XLT on Masakha. Results with XLM-R (X) and DeBERTa (D). 
}
\label{}
\end{table*}

\begin{table*}
\small
\setlength{\tabcolsep}{9.5pt}
\centering
\begin{tabular}{@{}llccccccccccc@{}}
\toprule
          &   & ar   & da   & de   & de-st & id   & it   & kk   & nl   & tr   & zh   & Avg  \\ \midrule
ZS        & X  & 71.5 & 85.6 & 80.8 & 43.9  & 86.8 & 88.2 & 80.8 & 88.8 & 81.5 & 57.4 & 76.5 \\ \midrule
\multicolumn{13}{c}{\textit{\textbf{Translate-Test: non-WA}}}                               \\ \midrule
\codec     & X & 79.0 & 81.9 & 86.1 & 60.4  & 84.8 & 88.4 & 83.0 & 86.5 & 83.6 & 67.0 & 80.1 \\
\codec     & D & 79.9 & 81.8 & 85.5 & 58.8  & 85.8 & 89.0 & 83.2 & 86.0 & 84.2 & 67.5 & 80.2 \\ \midrule
\multicolumn{13}{c}{\textit{\textbf{Translate-Test: WA}}}                                   \\ \midrule
\mbertwa & X & 79.1 & 76.2 & 85.2 & 60.2  & 79.1 & 87.9 & 75.3 & 87.3 & 78.1 & 80.1 & 78.8 \\
\mbertwa      & D & 79.3 & 75.9 & 84.4 & 58.4  & 79.9 & 88.6 & 75.1 & 86.5 & 78.9 & 80.0 & 78.7 \\
\labsewa       & X & 80.2 & 75.8 & 85.2 & 61.0  & 84.1 & 88.2 & 82.6 & 86.6 & 82.4 & 82.5 & 80.9 \\
\labsewa       & D & 80.7 & 75.5 & 84.5 & 59.3  & 85.0 & 88.9 & 82.7 & 86.0 & 83.3 & 82.4 & 80.8 \\
\nllbwa     & X & 81.0 & 81.0 & 87.4 & 61.0  & 87.0 & 88.5 & 82.4 & 87.8 & 83.2 & 83.2 & 82.2 \\
\nllbwa     & D & 81.4 & 80.7 & 86.7 & 59.3  & 87.9 & 89.2 & 82.4 & 86.9 & 84.1 & 83.1 & 82.2 \\
\bottomrule
\end{tabular}
\caption{Detailed main results for translation-based XLT on xSID. Results with XLM-R (X) and DeBERTa (D). 
}
\label{}
\end{table*}

\begin{table*}
\section{Detailed Results: Impact of Fine-Tuning}
\scriptsize
\setlength{\tabcolsep}{4.5pt}
\centering
\begin{tabular}{@{}llccccccccccccccccccc@{}}
\toprule
           &   & bam  & ewe  & fon  & hau  & ibo  & kin  & lug  & luo  & mos  & nya  & sna  & swa  & tsn  & twi  & wol  & xho  & yor  & zul  & Avg  \\ \midrule
\multicolumn{21}{c}{\textit{\textbf{Non-Fine-Tuned WAs}}}                                                                                           \\ \midrule
\mbertwa  & X & 46.0 & 76.9 & 57.9 & 70.1 & 75.5 & 64.9 & 83.0 & 71.8 & 43.5 & 77.9 & 63.3 & 79.8 & 80.6 & 70.9 & 53.1 & 50.0 & 58.1 & 60.3 & 65.8 \\
\mbertwa  & D & 46.0 & 77.0 & 58.6 & 70.6 & 76.9 & 65.6 & 82.8 & 73.2 & 43.6 & 78.4 & 63.7 & 80.5 & 81.2 & 71.7 & 53.5 & 50.0 & 57.7 & 60.2 & 66.2 \\
\labsewa   & X & 54.7 & 79.1 & 68.2 & 74.1 & 72.7 & 69.7 & 83.6 & 70.7 & 49.5 & 77.5 & 80.5 & 81.3 & 81.3 & 71.9 & 63.2 & 70.9 & 48.3 & 76.9 & 70.8 \\
\labsewa   & D & 54.1 & 79.1 & 68.6 & 74.6 & 74.0 & 70.2 & 83.3 & 72.0 & 49.3 & 78.2 & 81.0 & 82.0 & 81.6 & 73.0 & 63.6 & 71.1 & 48.2 & 77.0 & 71.2 \\
\nllbwa & X & 55.6 & 80.1 & 70.5 & 74.6 & 75.0 & 70.4 & 84.8 & 73.6 & 52.4 & 77.8 & 81.5 & 82.2 & 82.2 & 75.2 & 67.2 & 73.4 & 60.1 & 78.6 & 73.1 \\
\nllbwa & D & 55.4 & 80.1 & 71.2 & 75.1 & 76.6 & 71.0 & 84.6 & 75.0 & 52.4 & 78.5 & 82.1 & 82.8 & 82.5 & 75.9 & 68.0 & 73.5 & 59.9 & 78.8 & 73.5 \\ \midrule
\multicolumn{21}{c}{\textit{\textbf{Fine-Tuned WAs}}}                                                                                               \\ \midrule
\mbertwa  & X & 51.4 & 78.7 & 61.3 & 70.9 & 75.4 & 66.8 & 82.7 & 72.2 & 47.6 & 77.5 & 71.8 & 81.5 & 79.5 & 70.8 & 62.1 & 56.0 & 61.7 & 63.8 & 68.4 \\
\mbertwa  & D & 51.1 & 78.8 & 62.1 & 71.4 & 77.0 & 67.6 & 82.5 & 73.6 & 47.6 & 77.9 & 72.3 & 82.1 & 79.8 & 71.7 & 62.5 & 56.2 & 61.2 & 63.8 & 68.8 \\
\labsewa   & X & 54.4 & 79.9 & 69.7 & 74.7 & 75.2 & 70.8 & 84.4 & 72.6 & 53.1 & 78.6 & 81.7 & 83.0 & 80.0 & 71.2 & 64.9 & 73.2 & 55.4 & 78.7 & 72.3 \\
\labsewa   & D & 53.8 & 79.9 & 70.1 & 75.2 & 76.7 & 71.5 & 84.2 & 74.1 & 53.2 & 79.1 & 82.3 & 83.6 & 80.4 & 72.0 & 65.4 & 73.3 & 55.3 & 78.8 & 72.7 \\
\nllbwa & X & 56.8 & 80.8 & 72.8 & 74.9 & 75.8 & 71.0 & 84.8 & 74.7 & 54.0 & 78.8 & 82.2 & 82.3 & 82.2 & 75.1 & 68.6 & 73.8 & 62.8 & 79.0 & 73.9 \\
\nllbwa & D & 56.6 & 80.8 & 73.3 & 75.4 & 77.3 & 71.7 & 84.6 & 76.3 & 53.7 & 79.2 & 82.8 & 82.9 & 82.6 & 75.8 & 69.3 & 74.0 & 62.5 & 79.1 & 74.3 \\ \bottomrule
\end{tabular}
\caption{Impact of \wa fine-tuning on translation-based XLT on Masakha. Results with XLM-R (X) and DeBERTa (D). 
}
\label{}
\end{table*}

\begin{table*}
\small
\setlength{\tabcolsep}{9.5pt}
\centering
\begin{tabular}{@{}llccccccccccc@{}}
\toprule
      &   & ar   & da   & de   & de-st & id   & it   & kk   & nl   & tr   & zh   & Avg  \\ \midrule
\multicolumn{13}{c}{\textit{\textbf{Non-Fine-Tuned WAs}}}                               \\ \midrule
\mbertwa  & X & 74.2 & 75.8 & 84.6 & 58.9  & 76.0 & 85.3 & 59.3 & 85.7 & 69.2 & 73.6 & 74.1 \\
\mbertwa  & D & 74.8 & 75.5 & 83.8 & 56.9  & 76.4 & 86.0 & 59.7 & 85.2 & 69.7 & 73.3 & 74.1 \\
\labsewa   & X & 78.8 & 75.4 & 84.8 & 59.5  & 82.0 & 86.4 & 81.8 & 86.6 & 82.7 & 82.2 & 80.0 \\
\labsewa   & D & 79.2 & 75.2 & 84.1 & 57.9  & 83.0 & 87.2 & 81.8 & 85.9 & 83.6 & 82.1 & 80.0 \\
\nllbwa & X & 80.0 & 81.0 & 87.3 & 61.0  & 86.5 & 87.8 & 81.8 & 87.3 & 83.8 & 82.6 & 81.9 \\
\nllbwa & D & 80.3 & 80.7 & 86.6 & 59.4  & 87.5 & 88.5 & 81.8 & 86.6 & 84.6 & 82.6 & 81.8 \\ \midrule
\multicolumn{13}{c}{\textit{\textbf{Fine-Tuned WAs}}} \\ \midrule
\mbertwa & X & 79.1 & 76.2 & 85.2 & 60.2  & 79.1 & 87.9 & 75.3 & 87.3 & 78.1 & 80.1 & 78.8 \\
\mbertwa      & D & 79.3 & 75.9 & 84.4 & 58.4  & 79.9 & 88.6 & 75.1 & 86.5 & 78.9 & 80.0 & 78.7 \\
\labsewa       & X & 80.2 & 75.8 & 85.2 & 61.0  & 84.1 & 88.2 & 82.6 & 86.6 & 82.4 & 82.5 & 80.9 \\
\labsewa       & D & 80.7 & 75.5 & 84.5 & 59.3  & 85.0 & 88.9 & 82.7 & 86.0 & 83.3 & 82.4 & 80.8 \\
\nllbwa     & X & 81.0 & 81.0 & 87.4 & 61.0  & 87.0 & 88.5 & 82.4 & 87.8 & 83.2 & 83.2 & 82.2 \\
\nllbwa     & D & 81.4 & 80.7 & 86.7 & 59.3  & 87.9 & 89.2 & 82.4 & 86.9 & 84.1 & 83.1 & 82.2 \\ \bottomrule

\end{tabular}
\caption{Impact of \wa fine-tuning on translation-based XLT on xSID. Results with XLM-R (X) and DeBERTa (D). 
}
\label{}
\end{table*}

\begin{table*}
\section{Detailed Results: MT Model}
\small
\setlength{\tabcolsep}{6.8pt}
\centering
\begin{tabular}{@{}llcccccccccccc@{}}
\toprule
           &      & bam  & ewe  & hau   & ibo  & kin  & sna  & swa   & twi   & xho   & yor   & zul  & Avg  \\ \midrule
\mbertwa  & NLLB & 51.1 & 78.8 & 71.4  & 77.0 & 67.6 & 72.3 & 82.1  & 71.7  & 56.2  & 61.2  & 63.8 & 69.2 \\
\labsewa   & NLLB & 53.8 & 79.9 & 75.2  & 76.7 & 71.5 & 82.3 & 83.6  & 72.0  & 73.3  & 55.3  & 78.8 & 73.7 \\
\nllbwa & NLLB & 56.6 & 80.8 & 75.4  & 77.3 & 71.7 & 82.8 & 82.9  & 75.8  & 74.0  & 62.5  & 79.1 & 75.1 \\ \midrule
\mbertwa  & GT   & 55.4 & 78.9 & 71.9 & 79.4 & 68.1 & 75.2 & 84.1 & 73.5 & 59.0 & 65.1 & 66.1 & 70.6 \\
\labsewa   & GT   & 59.6 & 79.3 & 74.2 & 79.3 & 72.4 & 84.7 & 86.0 & 73.5 & 75.2 & 61.6 & 81.2 & 75.2 \\
\nllbwa & GT   & 61.5 & 79.9 & 74.2  & 80.4 & 72.6 & 84.8 & 86.11 & 77.13 & 75.73 & 66.73 & 81.3 & 76.4 \\ \bottomrule
\end{tabular}
\caption{Detailed results for translation-based \xlt on Masakha with translations obtained from different MT models---Google Translation (GT) and NLLB (NLLB). Results with DeBERTa. 
}
\label{}
\end{table*}

\begin{table*}
\small
\setlength{\tabcolsep}{8.5pt}
\centering
\begin{tabular}{@{}llccccccccccc@{}}
\toprule
           &      & ar   & da   & de   & de-st & id   & it   & kk   & nl   & tr   & zh   & Avg  \\ \midrule
\mbertwa  & NLLB & 79.3 & 75.9 & 84.4 & 58.4  & 79.9 & 88.6 & 75.1 & 86.5 & 78.9 & 80.0 & 78.7 \\
\labsewa   & NLLB & 80.7 & 75.5 & 84.5 & 59.3  & 85.0 & 88.9 & 82.7 & 86.0 & 83.3 & 82.4 & 80.8 \\
\nllbwa & NLLB & 81.4 & 80.7 & 86.7 & 59.3  & 87.9 & 89.2 & 82.4 & 86.9 & 84.1 & 83.1 & 82.2 \\ \midrule
\mbertwa  & GT   & 81.3 & 76.0 & 85.6 & 58.9  & 79.7 & 90.2 & 76.3 & 87.8 & 82.0 & 83.2 & 80.1 \\
\labsewa   & GT   & 81.7 & 76.6 & 85.3 & 58.8  & 85.3 & 90.1 & 85.1 & 87.1 & 84.4 & 86.4 & 82.1 \\
\nllbwa & GT   & 82.6 & 81.4 & 87.6 & 58.8  & 87.1 & 91.9 & 84.6 & 89.0 & 86.2 & 86.7 & 83.6 \\ \bottomrule
\end{tabular}
\caption{Detailed results for translation-based \xlt on xSID with translations obtained from different MT models---Google Translation (GT) and NLLB (NLLB). Results with DeBERTa. 
}
\label{}
\end{table*}

\begin{table*}[ht!]
\section{Detailed Results: Language Coverage}
\small
\setlength{\tabcolsep}{10.9pt}
\centering
\begin{tabular}{@{}lccccccccccc@{}}
\toprule
           & hau  & ibo  & kin  & nya  & sna  & swa  & wol  & xho  & yor  & zul  & Avg  \\ \midrule
\labsewa   & 75.2 & 76.7 & 71.5 & 79.1 & 82.3 & 83.6 & 65.4 & 73.3 & 55.3 & 78.8 & 74.1 \\
\nllbwa & 75.4 & 77.3 & 71.7 & 79.2 & 82.8 & 82.9 & 69.3 & 74.0 & 62.5 & 79.1 & 75.4 \\ \bottomrule
\end{tabular}
\caption{Detailed results for translation-based \xlt on Masakha only evaluating languages seen in the pretraining of both \was. Results with DeBERTa. 
}
\label{}
\end{table*}

\begin{table*}[ht!]
\small
\setlength{\tabcolsep}{12.8pt}
\centering
\begin{tabular}{@{}lcccccccccc@{}}
\toprule
           & ar   & da   & de   & id   & it   & kk   & nl   & tr   & zh   & Avg  \\ \midrule
\labsewa   & 80.7 & 75.5 & 84.5 & 85.0 & 88.9 & 82.7 & 86.0 & 83.3 & 82.4 & 83.2 \\
\nllbwa & 81.4 & 80.7 & 86.7 & 87.9 & 89.2 & 82.4 & 86.9 & 84.1 & 83.1 & 84.7 \\ \bottomrule
\end{tabular}
\caption{Detailed results for translation-based \xlt on xSID only evaluating languages seen in the pretraining of both \was. Results with DeBERTa. 
}
\label{}
\end{table*}

\begin{table*}
\section{Detailed Results: NLLB Model Size}
\scriptsize
\setlength{\tabcolsep}{4.3pt}
\centering
\begin{tabular}{@{}llccccccccccccccccccc@{}}
\toprule
           &      & bam  & ewe  & fon  & hau  & ibo  & kin  & lug  & luo  & mos  & nya  & sna  & swa  & tsn  & twi  & wol  & xho  & yor  & zul  & Avg  \\ \midrule
\nllbwa & 600M & 56.6 & 80.8 & 73.3 & 75.4 & 77.3 & 71.7 & 84.6 & 76.3 & 53.7 & 79.2 & 82.8 & 82.9 & 82.6 & 75.8 & 69.3 & 74.0 & 62.5 & 79.1 & 74.3 \\
\nllbwa & 3.3B & 57.1 & 80.7 & 74.0 & 75.2 & 77.3 & 71.8 & 84.6 & 76.3 & 54.1 & 79.6 & 82.7 & 83.3 & 82.5 & 75.8 & 69.5 & 74.0 & 62.8 & 79.2 & 74.5 \\ \bottomrule
\end{tabular}
\caption{Detailed results for translation-based XLT on Masakha with different sizes of NLLB as WA. Results with DeBERTa. 
}
\label{}
\end{table*}

\begin{table*}
\small
\setlength{\tabcolsep}{9pt}
\centering
\begin{tabular}{@{}llccccccccccc@{}}
\toprule
           &      & ar   & da   & de   & de-st & id   & it   & kk   & nl   & tr   & zh   & Avg  \\ \midrule
\nllbwa & 600M & 81.4 & 80.7 & 86.7 & 59.3  & 87.9 & 89.2 & 82.4 & 86.9 & 84.1 & 83.1 & 82.2 \\
\nllbwa & 3.3B & 80.8 & 76.0 & 86.2 & 59.3  & 82.5 & 89.4 & 83.5 & 86.8 & 86.3 & 83.7 & 81.4 \\ \bottomrule
\end{tabular}
\caption{Detailed results for translation-based XLT on xSID with different sizes of NLLB as WA. Results with DeBERTa. 
}
\label{}
\end{table*}

\end{document}